\documentclass[final,3p,times]{article}

\usepackage{amssymb}
\usepackage{PRIMEarxiv}

\usepackage{latexsym}
\usepackage[latin1]{inputenc}
\usepackage[T1]{fontenc}    
\usepackage{times}
\usepackage{amssymb}
\usepackage{algorithm}
\usepackage{algorithmic}
\usepackage{epsfig,shadow,amssymb,bm}
\usepackage{booktabs}
\usepackage{multirow}
\usepackage{graphicx}
\usepackage{tikz}
\usepackage{amsmath}
\usepackage{stackrel}
\usepackage{xurl}
\usepackage{hyperref}       
\usepackage{url}            
\usepackage{booktabs}       
\usepackage{amsfonts}       
\usepackage{nicefrac}       
\usepackage{microtype}      
\usepackage{lipsum}
\usepackage{fancyhdr}       
\usepackage{graphicx}

\usetikzlibrary{shapes,arrows}
\usetikzlibrary{positioning}
\usetikzlibrary{arrows,automata,decorations,shapes}
\tikzset{
  treenode/.style = {shape=rectangle, rounded corners,
                     draw, align=center,
                     top color=white, bottom color=blue!20},
  root/.style     = {treenode, font=\Large, bottom color=red!30},
  env/.style      = {treenode, font=\ttfamily\normalsize},
  dummy/.style    = {circle,draw}
}

\newcommand{\CASE}[1]{\STATE \textbf{case} #1\textbf{:} \begin{ALC@g}}
\newcommand{\ENDCASE}{\end{ALC@g}}

\newcommand{\DEFAULT}{\STATE \textbf{default:} \begin{ALC@g}}
\newcommand{\ENDDEFAULT}{\end{ALC@g}}
\newcommand{\DEFAULTLINE}[1]{\STATE \textbf{default:} }

\title{Multivariate feature ranking of gene expression data\thanks{An extended version of this paper has been published in \textit{IEEE Access} journal: F. Jiménez, G. Sánchez, J. Palma, L. Miralles-Pechuán and J. A. Botía, "Multivariate feature ranking with high-dimensional data for classification tasks," in \textit{IEEE Access}, 2022, doi: 10.1109/ACCESS.2022.3180773.}}

\author{Fernando Jim\'enez, Gracia Sánchez, Jos\'e Palma\\
	Artificial and Knowledge Engineering Group\\
	University of Murcia\\ 
	Spain \\
	\texttt{\{fernan,gracia,jtpalma\}@um.es} \\
	\And
	Luis Millares \\
	School of Computer Science\\
	Tecnological University Dublin\\ 
	Ireland \\
		\texttt{Luis.Miralles@TUDublin.ie} \\
	\And
	Juan Bot\'{\i}a  \\
	Intelligent System and Telematics Group\\ 
	University of Murcia\\
	Spain \\
	\texttt{juanbot@um.es} \\
}

\begin{document}

\maketitle

\begin{abstract}
	Gene expression datasets are usually of high dimensionality and therefore require efficient and effective methods for identifying the relative importance of their attributes. Due to the huge size of the search space of the possible solutions, the attribute subset evaluation feature selection methods tend to be not applicable, so in these scenarios feature ranking methods are used. Most of the feature ranking methods described in the literature are univariate methods, so they do not detect interactions between factors. In this paper we propose two new multivariate feature ranking methods based on pairwise correlation and pairwise consistency, which we have applied in three gene expression classification problems.
We statistically prove  that the  proposed methods outperform the state of the art feature ranking methods Clustering Variation,
Chi Squared,
Correlation,
Information Gain,
ReliefF and Significance, as well as feature selection methods of attribute subset evaluation based on correlation and consistency with multi-objective evolutionary search strategy.

\end{abstract}




%


\section{Introduction}

In the last decades, we have witnessed a substantial improvement in technologies for genome-wide gene expression (GE) analyses. Techniques such as GE arrays and RNA-Seq have allowed quantizing expression levels of thousands of genes over a set of biological samples in parallel, becoming an extraordinary and powerful tool in biomedical research. This is evidenced by more than  75,000 papers on gene expression that have been published in PubMed since 2020. All this research work has generated tons of data, most of them shared with the biomedical research community through well-known repositories such as Gene Expression Omnibus (GEO) \cite{edgar2002gene}, Expression Atlas EMBL-EBI \cite{papatheodorou2020expression}, GWAS Catalog \cite{buniello2019nhgri} and European Genome-Phenome Archive (EGA)  \cite{lappalainen2015european}. The analysis of gene expression data has allowed implementing new techniques for biomarker discovery (set of genes or networks with diagnostic and prognosis capabilities) \cite{perscheid2021integrative}, drug screening \cite{yang2020high} and the integration of metabolomics and multi-omics data \cite{metabo10050202}. 

One of the main tasks in analysing GE data is the discovery of genes affected by a given condition (e.g., disease, age, sex), also known as differentially expressed genes (DEG).  Traditionally, DEG techniques have been based on statistical methods which allow performing both univariate and multivariate analysis. This approach can be framed within what is known as \textit{filter-based} techniques for \textit{Feature Selection} (FS). The main disadvantage is that filter techniques only take into account characteristics inherent to the data, isolating the DEG discovery from the task to be solved (diagnosis, prognosis or clustering). 
The use of filter-based  FS techniques has increased the range of performance measures to be used for univariate and multivariate DEG discovery, allowing their integration with search strategies for subset evaluation. Filter methods have also been used successfully for \textit{Feature Ranking} (FR). 
FR methods assign 
a ranking or importance to each attribute, and they can also be treated as FS techniques
if a subset with the best attributes in the ranking, or those
above a certain threshold of importance, is selected.
Examples of  filter methods applied to gene expression data include \textit{mutual information} \cite{dabba2021gene}, \textit{information gain} \cite{zhang2020feature}, \textit{minimum redundancy - maximum relevance} \cite{baliarsingh2021sara} and \textit{symmetric uncertainty} \cite{kavitha2020score}.  An exhaustive analysis of filter techniques for FS in GE data can be found in \cite{lazar2012survey}. 

In the last decade, there is a growing interest in applying \textit{machine learning} techniques to GE analysis.  The use of machine learning techniques has made possible the application of new approaches for DEG discovery. This is the case of \textit{wrapper} FS techniques which have been successfully applied, mainly focused on multivariate FS. In contrast to a filter-based method, wrapper FS methods build a predictive model to evaluate attributes, either individually or attribute subsets. Obviously, wrapper FS methods require more computational time than filter-based methods since a predictive model has to be fitted for each candidate subset. 
The main advantage of this approach relies on the fact that  attribute evaluation is task-oriented. In other words, attributes are evaluated according to their predictive power by improving predictive model performance. For example, in \cite{rustam2020comparison} \textit{support vector machines} have been used with \textit{recursive feature elimination} as the search strategy whilst  \cite{kalaimani2020novel} used a \textit{best first} search. Apart from their high computation cost, another disadvantage of wrapper methods is that predictive model overfitting can affect the solution quality. To overcome this problem,  there is a growing interest in the use of hybrid techniques, which try to integrate two or more different FS methods, finding informative genes and reducing the computational cost \cite{panigrahi2021survey}.  Moreover, different subsets search strategies have been proposed to improve the efficiency of FS methods, such as \textit{metaheuristic} techniques \cite{jimenez2017multi, 6256452, Elg2020}. These strategies do not guarantee to find the optimal subsets but reach acceptable solutions in terms of a trade-off between optimality and computing effort. Their main advantage is that they avoid being trapped in a local minimum  (maximum) as can be the case of deterministic techniques, due to their global optimization approach. This, together with the fact that they allow defining more than one objective and several constraints to the optimization problem, has meant that they have been successfully applied to FS problems such us DEG discovery \cite{shukla2020study}.

Since gene expression  datasets usually, as previously said, contain thousands of attributes, subset evaluation FS methods are often inefficient in this scenario. Search space of this FS problems is $O(2^n)$, where $ n $ is the number of attributes, and heuristics and metaheuristics can only find satisfactory solutions when very long computation times are used, even if the FS method is of type filter. The computation time required increases even more when we use wrapper feature selection methods, which can become impractical. FR methods, which evaluate attributes individually instead of evaluating attribute subsets, are probably the most viable alternative for this type of scenario. However, existing FR methods are univariate methods, with the exception of \textit{ReliefF} \cite{kononenko1994estimating} which is a multivariate FR method. Univariate FR methods  evaluate the attributes in a ``myopic'' way. That is, not considering interdependencies or interactions between the attributes.
If each attribute is evaluated without considering the rest of the attributes, we will probably obtain poor results, because the attribute interdependencies are not being considered. In this paper, we propose two novel multivariate FR methods based on pairwise correlation and pairwise consistency respectively.
The proposed FR methods are compared with a wide range of FR methods (including the multivariate \textit{ReliefF} method) and with subset evaluation FS methods based on  correlation and  consistency with a multi-objective evolutionary search strategy. 

The paper has been organized as follows: section \ref{background} shows the state-of-the-art of FS methods for GE data;
section \ref{PCONAE-PCORAE} describes the novel multivariate FR methods proposed in this paper; section \ref{Experiments} describes the datasets and the performed experiments; section \ref{AnaResDis} analyses the obtained results;  finally, section \ref{Conclusions} presents conclusions and outlines future work.

\section{Related works}
\label{background}

In \cite{sambhe2021multi}, the authors try to classify 10 different types of proteins, ranging from 6 to 3,822 nucleotides. For each sample, k-mers of length 6 have been used as features and their frequency was normalized using Term Frequency Inverse Document Frequency (TF-IDF).  In the work, a comparison of the single-objective and multi-objective approaches for FS and classification is performed. In the single-objective approach,  two FR techniques and principal component analysis (PCA) have been used: analysis of variance (ANOVA) and Chi-Square. The features selected have been used for building twelve classification to find out which combination presents better performance. In the multi-objective approach, a metaheuristic wrapper based technique is used to simultaneously find the best combination of feature subsets and classification techniques. In this case, the metaheuristics techniques used are NSGA-II multi-objective evolutionary and Multi-dimensional Archive of Phenotypic Elites (MAP-Elites) algorithms. The results reveal that the two approaches produce good outcomes, being the computational cost of models based on FR techniques substantially lower. However, metaheuristics based approaches perform a better exploration of the search spaces. Specifically, NSGA-II showed a tendency in selecting fewer features whereas MAP-Elites presented a wider Pareto front achieving competitive results. 

An interesting benchmark of ranker methods in high-dimensional gene expression applied to 11 survival datasets is presented in \cite{bommert2021benchmark}. Specifically, 14 ranker based filter methods have been applied.  Among them, we can find those based on variance, correlation, Cox score and MI. Ranker based filter methods are compared with regularized Cox proportional hazards model using all features. The conclusions drawn state that models, obtained after applying one of the filters, achieve better predictive power than the baseline model. Furthermore, models with better predictive power only select a small number of features.  One of the most interesting characteristics of ranker methods is that they provided, as output, a ranking of the features based on a concrete univariate measure. This facilitates the aggregation of different FR methods as part of an ensemble-based strategy. An example of this approach is presented in \cite{10.1007/978-3-030-86653-2_12}. In this work, different versions of the microarray data set are generated by bootstrapping the original set. In each bootstrapped bag a different ranker technique is applied. The final ranking is the result of the average ranking of each feature across all the partial rankings. This approach has also been successfully applied in \cite{PANT2019103254} for stomach cancer biomarker identification.  In this case, four different ranking techniques have been applied: Conditional Mutual Information Maximisation (CMIM) \cite{fleuret2004fast}, Double Input Symmetrical Relevance (DISR) \cite{meyer2006use}, Interaction Capping (ICAP) \cite{jakulin2005machine} and Conditional Informative Feature Extraction (CIFE) \cite{lin2006conditional}. A novel technique, Weighted Ensemble of Ranks (WER$_j$) is proposed to aggregate individual rankings. Then, among the top 100 genes, only those common to all different are selected. The proposed method suggests that the selected genes show better performance accuracy when multiple clinical outcomes are considered. 

Despite being MI  one of the most used FS ranking measure, obtaining a reliable estimation of MI on high-dimensional low-sample datasets remains a challenge. To this end, in \cite{sharmin2019simultaneous}, a novel Joint Bias Mutual Information (JBMI) is first presented, together with modified Discretization and Selection of feature based on MI (mDSM). The authors also demonstrate that MI follows a $\chi^2$ distribution, making it possible to design an FS technique that, simultaneously to FS, selects the best discretization of selected features using the $\chi^2$ criteria. These findings have been used in  \cite{haque2021use}, where a novel Mutual information-based Gene Selection (MGS) is presented. In order not to lose relevant genes, mDSM is applied in a Leave One Out Cross Validation scheme, resulting in different rankings of relevant genes. Then two ranking criteria have been applied to aggregate different rankings namely MGS frequency-based ranking (MGS$_f$ )and MGS Random Forest based ranking (MGS$_rf$). An evaluation of the proposed techniques has been performed over different GE datasets. Results proofed that both  (MGS$_f$ ) and (MGS$_rf$) outperforms existing techniques in balanced and imbalanced datasets.

An interesting approach using rankers based on Autoencoders (AEs)  is presented in \cite{macias2020autoencoded}. First, an AE is applied for a non-linear fusion of and summarization of the original characteristics. Then,  the FR technique ANOVA with FDR correction is used to retain the most relevant features which have been used in different ML classification techniques. The conclusions showed that the combination of AE and FR provided better performance. It could be argued that the use of AE does not allow the extraction of biomedical related conclusions since information about original features has been lost in the AE fusion process. However, it is worthy to mention here that a methodology to calculate gene weights for genes of a set of AE features is also presented.

\section{Materials and methods} 
\label{PCONAE-PCORAE}

Let $D=\{I_1,\ldots,I_w\}$ be a dataset with $w$ instances. Each instance $I_p=(a_1^p,\ldots,a_n^p,c_p)$, $p=1\ldots,w$, has $n$ input attributes of any type,
and one output attribute $c_p\in\{1,\ldots,s\}$,  where $s$ is the number of output classes.
We assume that at least one instance exists for each output class. Below we describe two new multivariate FR methods based on pairwise correlation and pairwise consistency respectively, which are proposed in this paper for GE data.

\subsection{Multivariate feature ranking based on pairwise correlation}

We propose the FR method called  \textit{Pairwise Correlation}, which is inspired in the \textit{correlation-based feature selection} (CFS) method. The CFS algorithm was developed by Mark A. Hall from the University of Waikato in Hamilton (New Zealand) throughout his doctoral thesis  \cite{hall1999correlation}. This same university is well-known in the world of data science for developing the free software application \textit{Weka} (\textit{Waikato Environment for Knowledge Analysis} \cite{Weka}), in which Mark A. Hall has an important role as Honorary Research Associate.
CFS algorithm has the advantage over other FS methods that it  generates more accurate models and reduces the number of attributes selected  by half in most cases. This concept is explained years later in a summary paper \cite{hall2000correlation}. The CFS  method evaluates sets of attributes instead of doing so individually. To determine the goodness of each set, the CFS algorithm evaluates how well each attribute is able to predict the class as well as the similarity degree between the attributes. In such a way that the feature sets correlated with the class and with features poorly correlated with each other obtain the higher scores. 
CFS method uses the function ${\Phi}_D(\mathcal{S})$ to measure the quality of a subset $\mathcal{S}$ of $k$ attributes in a dataset $D$, $1\leq k\leq n$, defined as follows:

\begin{equation}\label{eq:Cfs}
\Phi_D(\mathcal{S}) = \frac{k \cdot \sigma_{D}^c}{\sqrt {k+ k\cdot (k-1) \cdot \sigma_{D}^f}}
\end{equation}

\noindent where $\sigma_{D}^c$ is
the mean of the correlations between each feature in $\mathcal{S}$ and the class attribute, and $\sigma_{D}^f$ is the average correlation between each of the $\binom{k}{2}$ possible feature pairs
in $\mathcal{S}$. In other words, the numerator indicates the predictive degree of a set of variables while the denominator indicates the redundancy between the variables. CFS method requires discretizing the values (usually with Fayyad and Irani method \cite{fayyad1993multi}). CFS applies the \textit{symmetrical uncertainty} method \cite{press1988numerical} to measure the degree of similarity for discrete values. 

The proposed FR method \textit{Pairwise Correlation} evaluates an attribute $i\in\{1,\ldots,n\}$ by using the following function $\Phi_D^A$: 
\begin{equation}
	\Phi_D^A(i)=\frac{1}{n-1}\cdot\displaystyle\sum_{\substack{j\in\{1,\ldots, n\} \\ j\neq i}}\Phi_D(\{i,j\})
\end{equation}
where $\Phi_D(\{i,j\})$ is the merit (eq. (\ref{eq:Cfs})) of the subset formed by attributes $i$ and $j$, for all $j=1,\ldots, n$, with $j\neq i$. That is, the merit $\Phi_D^A(i)$ of an attribute $i$ is the mean of the merits $\Phi_D$ of the attribute subsets formed by $i$ and each of the other attributes. Attributes with low correlation to other attributes and highly correlated with the class are preferred.
\textit{Pairwise Correlation} is a (filter) multivariate FR method since the evaluation of each attribute takes into account all the other attributes together with the class, thus considering the interactions between the attributes.

\subsection{Multivariate feature ranking based on pairwise consistency}

Similarly, we propose the FR method called  \textit{Pairwise Consistency}, which uses the \textit{consistency} metric for attribute subsets introduced by Liu and Setiono \cite{liu1996probabilistic}.
The intuition behind the consistency measure is to find attributes that divide the dataset into parts with a highly predominant class. This measure has been explained by H. Almuallim et al. in 1991 \cite{almuallim1991learning} and by H. Liu and R. Setiono in 1996 \cite{liu1996probabilistic}, but the research of Dash and H. Liu \cite{dash2003consistency} gives a more complete perspective. 

According to \cite{dash2003consistency}, a group of features is \textit{inconsistent} when two or more instances have the same values but different labels. For example, the instances $(1,2,2,a)$ and $(1,2,2,b)$, where $a$ and $b$ represent the class, are inconsistent. The consistency measure is then defined by the \textit{inconsistency rate}. The \textit{inconsistency rate}, $I_D(\mathcal{S})$, of a attribute subset $\mathcal{S}$ in a dataset $D$ is calculated as the sum of all the inconsistency counts for all the patterns divided by the total number of instances in $D$. The inconsistency count for a given pattern (the values of the selected features without the class) is calculated as the total number of same patterns in the dataset minus the number of instances of the majority class of the pattern. For example, for the pattern $(1,2,2)$, if there are $36$ elements of class $a$,  $6$ for class $b$, and $5$  for  class $c$, the inconsistency count would be $(36+6+5) - 36 = 11$. Obviously, the less inconsistent the subset is, the greater the  consistency of an attribute subset. The consistency of any subset can never be lower than that of the full set of attributes. The usual practice is to use this subset evaluator in conjunction with a search strategy which looks for the smallest subset with consistency equal to that of the full set of attributes. Consistency measure can work
when data has discrete valued features. Any continuous feature should be first discretized
using some discretization method \cite{fayyad1993multi}. 

The proposed FR method \textit{Pairwise Consistency} evaluates an attribute $i\in\{1,\ldots,n\}$ by using the following function $\Psi_D^A$:

\begin{equation}
\Psi_D^A(i)=\frac{1}{n-1}\cdot\displaystyle\sum_{\substack{j\in\{1,\ldots, n\} \\ j\neq i}}\Psi_D(\{i,j\})
\end{equation}
where $\Psi_D(\{i,j\})=1-I_D(\{i,j\})$ is the \textit{consistency rate} of the subset formed by the attributes $i$ and $j$, for all $j=1,\ldots, n$, with $j\neq i$. That is, the merit $\Psi_D^A(i)$ of an attribute $i$ is the mean of the consistency rates of the attribute $i$ and each of the other attributes.  \textit{Pairwise Consistency} is also a (filter) multivariate FR method and therefore considers interactions between factors.

\section{Experiments and results}
\label{Experiments}
In this section we describe the gene expression datasets used in this paper (section \ref{sec:datasets}) as well as the experiments performed and their results. The proposed FR methods are compared to 6 other FR methods using a set of 8 classification algorithms and the  performance metric of percent correct. Statistical tests are performed to detect statistically significant differences between FR methods and to establish a win-loss ranking. Furthermore, the proposed FR methods have been compared with other filter FS methods for attribute subset evaluation that use metaheuristic search strategies. The results of these two groups of experiments are shown in sections \ref{sec:FR-comparison} and \ref{sec:SE-comparison} respectively.

\subsection{Gene expression datasets}
\label{sec:datasets}

Two gene expression datasets are described below for classification tasks, named \textit{gene expression cancer RNA-Seq} and \textit{GTEx RNA expression} datasets. From the second dataset we obtain, in turn, two classification problems called \textit{brain tissue GTEx RNA expression} and \textit{brain age GTEx RNA expression}.

\paragraph{Gene expression cancer RNA-Seq dataset}
The \textit{gene expression cancer RNA-Seq} dataset\footnote{\url{https://archive.ics.uci.edu/ml/datasets/gene+expression+cancer+RNA-Seq}}  is part of the RNA-Seq (HiSeq) PANCAN dataset. The original dataset\footnote{\url{https://www.synapse.org/\#!Synapse:syn4301332}} is maintained by the cancer genome atlas pan-cancer analysis project \cite{CancerGenomeAtlasResearchNetwork:2013:Nat-Genet:24071849}. This dataset is a random extraction of gene expressions of patients having different types of tumor, containing 801 instances and 20531 input attributes. Attributes of each instance are RNA-Seq gene expression levels measured by illumina HiSeq platform. A dummy name (\textit{gene\_xxxxx}) is given to each attribute. Attributes are ordered consistently with the original submission. It is possible to obtain the complete list of names by accessing the project platform. The output attribute has 5 classes corresponding to the 5 tumor types shown in Table \ref{tab:cancer-classes} along with the number of instances of each class.

\begin{table}[!htpb]
	\centering
	\begin{tabular}{cc}\hline		
		\textbf{\textit{Tumor}} & 
		\textbf{\textit{Number of instances}} \\\hline
		BRCA & 300  \\
		COAD & 78  \\ 
		KIRC & 146  \\
		LUAD & 141  \\
		PRAD & 136  \\			
		\hline			
	\end{tabular}
	\caption{Classes and their number of instances of the gene expression cancer RNA-Seq dataset.}
	\label{tab:cancer-classes}
\end{table}

\paragraph{GTEx RNA expression dataset} GTEx \cite{GTEx} is an international consortium devoted to sequencing multiple parts of the human body, including 13 different brain areas, the main organs, e.g., lung, liver, heart, sking.  Humans are all control subjects from a variety of ages and sex. To date, the GTEx transcriptomic resource is the biggest repository of human tissue RNA and DNA sequencing. We downloaded TPM (Transcript per million) values from the global expression matrix of GTEx RNA expression V7 and biological covariates of interest like sex, age and sample tissue. We filter out all non-brain tissue samples and kept only those genes expressed with a minimum of 0.1 TMP values over 80\% of the samples, within each tissue. Then separately for each brain tissue, we regressed out of the expression, RIN, age and sex to avoid both technical and biological biases for this specific experiment. The resultant expression values are available in the form of a R package downloadable from GitHub\footnote{\url{https://github.com/juanbot/CoExpGTExV7}}. The resulting dataset consists of 17863 input attributes and 1529 instances. We created two classification problems out of this dataset. The classes for the \textit{brain tissue GTEx RNA expression} classification problem, shown in Table \ref{tab:tissue-classes} along with the number of instances of each class,  were the tissue corresponding to the specific individual sample.  The classes for the \textit{brain age GTEx RNA expression} classification problem, shown in  Table \ref{tab:age-classes}, corresponds with 6 ranges of age for the individual.

\begin{table}[!htpb]
	\centering
	\begin{tabular}{cc}\hline		
		\textbf{\textit{Brain tissue}} & 
		\textbf{\textit{Number of instances}} \\\hline
		BrainAmygdala & 97  \\ 
		BrainAnteriorCingulateCortex  & 121   \\
		BrainCaudateBG & 157   \\
		BrainCerebellarHemisphere & 134   \\
		BrainCerebellum & 173   \\
		BrainCortex & 158   \\
		BrainHippocampus & 123   \\
		BrainHypothalamus & 120   \\
		BrainNucleusAccumbensBG & 146  \\
		BrainPutamenBG & 123   \\
		BrainSpinalCordC1 & 90   \\
		BrainSubstantiaNigra & 87 	  \\
		\hline			
	\end{tabular}
	\caption{Classes and their number of instances of the brain tissue GTEx RNA expression classification problem.}
	\label{tab:tissue-classes}
\end{table}

\begin{table}[!htpb]
	\centering
	\begin{tabular}{cc}\hline		
		\textbf{\textit{Brain age}} & 
		\textbf{\textit{Number of instances}} \\\hline
		20-29 & 59 \\
		30-39 & 35 \\
		40-49 & 165 \\
		50-59 & 478 \\
		60-69 & 705 \\
		70-79 & 87 \\
		\hline			
	\end{tabular}
	\caption{Classes and their number of instances of the brain age GTEx RNA expression classification problem.}
	\label{tab:age-classes}
\end{table}

\subsection{Comparison with other feature ranking methods}
\label{sec:FR-comparison}
Our two methods \textit{Pairwise Correlation} and \textit{Pairwise Consistency} have been compared
with 6 other well-known FR methods that use different information measures. These FR methods are \textit{Clustering Variation} \cite{fong2014novel},
 \textit{Chi Squared} \cite{galavotti2000experiments},
 \textit{Correlation} \cite{freedman2007statistics},  
 \textit{Information Gain} \cite{Karegowda2010},  
 \textit{ReliefF} \cite{kononenko1994estimating} and 
 \textit{Significance} \cite{Ahm2005}. To compare the FR methods, the reduced datasets containing the $q$ best attributes obtained with each FR method have been used to construct classifiers of different nature, for $q=3,\log_2(n),50,100$. The classification algorithms used were \textit{naive bayes} \cite{DBLP:reference/ml/Webb17f}, \textit{multilayer perceptron} \cite{alma991043683361603276}, \textit{support vector machine} \cite{Cortes95support-vectornetworks}, \textit{k-NN} \cite{Cover:2006:NNP:2263261.2267456}, \textit{ripper} \cite{Cohen95fasteffective}, \textit{C4.5} \cite{j48}, \textit{random forest} \cite{Breiman2001} and \textit{zeroR} \cite{brownleeZeroR}. 
 Each pair (reduced dataset, classification algorithm) has been evaluated using 10-fold cross-validation, repeated 10 times. Therefore, for each pair (reduced dataset, classification algorithm), 100 classifiers have been constructed, which have been evaluated with the metric of \textit{percent correct} and the mean has been calculated. Tables \ref{tab:PECancer-q3} to \ref{tab:PEAGE-q100} show these results, for each value of $ q $, in the three classification problems considered. Table \ref{tab:PETotal} shows, as a summary, the number of times that each FR method has obtained the best  evaluation for some value of $q$, in each of the classification problems and the total. The FR methods were ordered according to this score and the rank of each FR method is shown. The FR methods in the first two positions in the ranking are marked in bold.

\begin{table}[!htpb]
	\resizebox{\textwidth}{!}{\begin{tabular}{cccccc}\hline		
\textbf{\textit{Method}} & \textbf{\textit{Cancer RNA-Seq}}
& \textbf{\textit{Brain tissue GTEx RNA}}
& \textbf{\textit{Brain age GTEx RNA}}
& \textbf{\textit{Total}}
& \textbf{\textit{Rank}}
\\ \hline
\textit{Clustering Variation}   & 0 & 0 & 0 & 0 & 6 \\
\textit{Chi Squared}  & 1 & 0 & 0 & 1 & 5 \\
\textit{Correlation}   & 0 & 0 & 0 & 0 & 6 \\
\textit{Information Gain}   & 3 & 1 & 0 & 4 & 3 \\
\textit{ReliefF}  & 1 & 0 & 0 & 1 & 5 \\
\textit{Significance}  & 1 & 0 & 2 & 3 & 4 \\
\textbf{\textit{Pairwise Correlation}}  & 3 & 2 & 1 & \textbf{6} & \textbf{1} \\
\textbf{\textit{Pairwise Consistency}}  & 2 & 2 & 1 & \textbf{5} & \textbf{2} \\
			
			\hline			
		\end{tabular}}
		\caption{Number of times each method was the best in each problem, along with the total number of times and the final rank for each method.}
		\label{tab:PETotal}
	\end{table}	
							
We have performed statistical tests to detect statistically significant differences between the compared FR methods. A \textit{paired t-test} has been performed establishing as baseline the reduced dataset obtained with each of the FR methods, for each value of $q$ and in each classification problem. In this way, each FR method has been statistically compared with all the others. Then, the times that each FR method has won (wins) and the times that it has lost (losses) are obtained. Finally, the FR methods are ordered according to the difference between wins and losses. Tables \ref{tab:cancer} to \ref{tab:brainage} show the results of the statistical tests for each value of $q$ and for each classification problem. In these tables, the FR methods with the greatest difference between wins and losses in each value of $q$ have been marked in bold.  Finally, Table \ref{tab:PETotal} shows the final ranking of the FR methods taking into account all the values of $q$ and all the classification problems. The top 2 FR methods in the ranking have been marked in bold.

\begin{table}[!htpb]
	\resizebox{\textwidth}{!}{\begin{tabular}{c|cc|cc|cc|cc|}\cline{2-9}
			
				& \multicolumn{2}{c|}{\textbf{\textit{Cancer RNA-Seq}}}
				& \multicolumn{2}{c|}{\textbf{\textit{Brain tissue GTEx RNA}}}
				& \multicolumn{2}{c|}{\textbf{\textit{Brain age GTEx RNA}}}
				& \multicolumn{2}{c|}{\textbf{\textit{Total}}}
				\\ \hline
			\multicolumn{1}{|c|}{\textbf{\textit{Method}}} & 
			\textbf{\textit{Difference}} &
			\textbf{\textit{Rank}} & \textbf{\textit{Difference}}  &
			\textbf{\textit{Rank}}  &
			\textbf{\textit{Difference}}  &
			\textbf{\textit{Rank}}  &
			\textbf{\textit{Difference}}  &
			 \textbf{\textit{Rank}}\\\hline
			\multicolumn{1}{|c|}{{\textit{Clustering Variation}}}   & -147 & 8 & -86 & 6 & -57 & 6 & -250 & 8 \\
			\multicolumn{1}{|c|}{{\textit{Chi Squared}}}  & 41 & 4 & 63 & 4 & 30 & 3 & 134 & 4 \\
			\multicolumn{1}{|c|}{{\textit{Correlation}}}  & -37 & 7 & -132 & 8 & -69 & 7 & -238 & 7 \\
			\multicolumn{1}{|c|}{{\textit{Information Gain}}}  & 45 & 3 & 96 & 2 & 29 & 4 & 170 & 3 \\
			\multicolumn{1}{|c|}{{\textit{ReliefF}}}   & 6 & 6 & -116 & 7 & -49 & 5 & -159 & 6 \\
			\multicolumn{1}{|c|}{{\textit{Significance}}}  & 35 & 5 & 8 & 5 & 40 & 2 & 83 & 5 \\
			\multicolumn{1}{|c|}{{\textbf{\textit{Pairwise Correlation}}}}  & 54 & 1  & 99 & 1 & 29 & 4 & \textbf{182} & \textbf{1} \\
			\multicolumn{1}{|c|}{{\textbf{\textit{Pairwise Consistency}}}} & 52 & 2 & 72 & 3 & 47 & 1 & \textbf{171} & \textbf{2} \\
			\hline			
		\end{tabular}}
		\caption{Ranking of the comparison with attribute evaluation feature selection methods.}
		\label{tab:AErank}
	\end{table}

\subsection{Comparison with attribute subset evaluation feature selection methods}
\label{sec:SE-comparison}

This section compares the FR \textit{Pairwise Correlation} and \textit{Pairwise Consistency} methods with FS methods that use the correlation and consistency filters but for attribute subset evaluation instead of attribute evaluation. These FS methods  require a strategy to search for candidate subsets of attributes in a search space $ O (2 ^ n) $.
We have used a multi-objective evolutionary search strategy \cite{Jim15, icedm2016, jimenez2017multi, Jimenez2019, Jimnez2019MultiobjectiveEF}, in particular the NSGA-II algorithm \cite{Deb02}, with which the merit of the attribute subsets is maximized and its cardinality is minimized. These FS methods are called \textit{MOEA-CFS} and \textit{MOEA-Consistency} in this paper. In the comparisons, we have used $ q = 100 $ and $ q = 200 $ for the FR methods \textit{Pairwise Correlation} and \textit{Pairwise Consistency}.
Again, each pair (reduced dataset, classification algorithm) has been evaluated using 10-fold cross-validation, repeated 10 times.
\textit{MOEA-CFS} and \textit{MOEA-Consistency} have been run with population size of 100 and 1000 generations (100000 evaluations of the objective function).
Tables \ref{tab:PECancer-SE} to \ref{tab:PEage-SE} show the average evaluations with the metric of \textit{percent correct}. Table \ref{tab:PETotal-SE} shows the number of times that each method has obtained the best  evaluation in each of the classification problems and the total, marking in bold the two best methods. Table \ref{tab:SEtest}  shows the results of the statistical tests  and the wins-losses ranking, marking in bold the best method in each classification problem.
Table \ref{tab:SErank} shows the final ranking of the methods taking into account all the classification problems, marking in bold the top 2  methods.

\begin{table}[!htpb]
	\resizebox{\textwidth}{!}{\begin{tabular}{cccccc}\hline		
			\textbf{\textit{Method}} & \textbf{\textit{Cancer RNA-Seq}}
			& \textbf{\textit{Brain tissue GTEx RNA}}
			& \textbf{\textit{Brain age GTEx RNA}}
			& \textbf{\textit{Total}}
			& \textbf{\textit{Rank}}
			\\ \hline
					{{\textit{MOEA-CFS}}}  & 0 & 1 & 0 & 1 & 2 \\
					{{\textit{MOEA-Consistency}}} & 0  & 0 & 0 & 0 &  3\\
					\textbf{{{\textit{Pairwise Correlation ($ q=100 $)}}}}& 2 & 0 & 0 & \textbf{2} & \textbf{1} \\
					\textbf{{{\textit{Pairwise Consistency ($ q=100 $)}}}}& 1 & 0 & 1 & \textbf{2} & \textbf{1} \\
					{{\textit{Pairwise Correlation ($ q=200 $)}}}& 0 & 0 & 0 & 0 & 3 \\
					{{\textit{Pairwise Consistency ($ q=200 $)}}}& 1 & 0 & 0 & 1 & 2 \\	
			\hline			
		\end{tabular}}
		\caption{Number of times each method was the best in each problem, along with the total number of times and the final rank for each method.}
		\label{tab:PETotal-SE}
	\end{table}

\begin{table}[!htpb]
	\resizebox{\textwidth}{!}{\begin{tabular}{c|cc|cc|cc|cc|}\cline{2-9}
			
			& \multicolumn{2}{c|}{\textbf{\textit{Cancer RNA-Seq}}}
			& \multicolumn{2}{c|}{\textbf{\textit{Brain tissue GTEx RNA}}}
			& \multicolumn{2}{c|}{\textbf{\textit{Brain age GTEx RNA}}}
			& \multicolumn{2}{c|}{\textbf{\textit{Total}}}
			\\ \hline
			\multicolumn{1}{|c|}{\textbf{\textit{Method}}} & 
			\textbf{\textit{Difference}} &
			\textbf{\textit{Rank}} & \textbf{\textit{Difference}}  &
			\textbf{\textit{Rank}}  &
			\textbf{\textit{Difference}}  &
			\textbf{\textit{Rank}}  &
			\textbf{\textit{Difference}}  &
			\textbf{\textit{Rank}}\\\hline
			\multicolumn{1}{|c|}{{\textit{MOEA-CFS}}}  & 7 & 1 & 5 & 3 & -4 & 5 & 8 & 4 \\
			\multicolumn{1}{|c|}{{\textit{MOEA-Consistency}}}  & -35 & 2 & -27 & 6 & 1 & 2 & -61 & 6 \\
			\multicolumn{1}{|c|}{{{\textit{Pairwise Correlation ($ q=100 $)}}}} & 7 & 1 & 1 & 4 & 3 & 1 & 11 & 3 \\
			\multicolumn{1}{|c|}{\textbf{{\textit{Pairwise Consistency ($ q=100 $)}}}} & 7 & 1 & 16 & 1 & 3 & 1 & \textbf{26} & \textbf{1} \\
			\multicolumn{1}{|c|}{{\textit{Pairwise Correlation ($ q=200 $)}}}  & 7 & 1 & -3 & 5 & -1 & 3 & 3 & 5 \\
			\multicolumn{1}{|c|}{{\textbf{\textit{Pairwise Consistency ($ q=200 $)}}}} & 7 & 1 & 8 & 2 & -2 & 4 & \textbf{13} & \textbf{2} \\
			\hline			
		\end{tabular}}
		\caption{Ranking of the comparison with attribute subset evaluation feature selection methods.}
		\label{tab:SErank}
	\end{table}	
	
%

\section{Analysis of results and discussion}
\label{AnaResDis}

Tables \ref{tab:PETotal}, \ref{tab:AErank} \ref{tab:PETotal-SE} and \ref{tab:SErank} allow us to analyze the results clearly, since they compile the summaries of all the experiments carried out in the three classification problems and in the total, both in the evaluation of the FR and FS methods and the results of statistical tests. The following statements can be derived from the results obtained:

\begin{itemize}
	\item The results indicate that the multivariate FR methods proposed in this paper statistically outperform the rest of the compared FR methods, both univariate and multivariate.
	\item If we compare the univariate FR method \textit{Correlation} with the multivariate FR method \textit{Pairwise Correlation} proposed in this paper, the results are clearly favourable to \textit{Pairwise Correlation}, although both FR methods are based on the correlation metric.
	\item The \textit{ReliefF} method, even being a multivariate method, has not shown good performance in the tested gene expression classification problems, even lower than  other univariate methods such as \textit{Chi Squared}, \textit{Information Gain} or \textit{Significance}.
	\item The proposed FR methods statistically outperforms multivariate FS methods of attribute subset evaluation based on correlation and consistency, with powerful search strategies, such as multi-objective evolutionary algorithms.
	This is due to the enormous search space that occurs with the gene expression datasets considered in this paper. For example, for the \textit{cancer RNA-Seq} dataset, there are $ 2 ^ {20531} = 2.8\text{e+}6180$ candidate subsets of attributes. This makes search strategies require a lot of execution time to obtain satisfactory solutions in these types of problems, which can be prohibitive in some cases.

\end{itemize}

\section{Conclusions and future work}
\label{Conclusions}

In this paper we have presented two new feature ranking methods that are especially appropriate for high-dimen\-sional datasets, such as gene expression datasets. These methods, which we have called \textit{Pairwise Correlation} and \textit{Pairwise Consistency}, are filter methods based on correlation and consistency respectively that evaluate each attribute by computing the average merit of all pairs of attributes formed by the attribute and each of the others, therefore being multivariate methods. Like any feature ranking method, \textit{Pairwise Correlation} and \textit{Pairwise Consistency} can also be used for feature selection.

We have compared the \textit{Pairwise Correlation} and \textit{Pairwise Consistency} methods with six feature ranking methods well known in the literature, and also with two feature selection methods of attribute subset evaluation based on correlation and consistency with multi-objective evolutionary search strategy.  Three classification problems have been considered for gene expression data: \textit{cancer RNA-Seq}, \textit{brain tissue GTEx RNA} and \textit{brain age GTEx RNA}. For comparisons, we used eight classification algorithms of different nature evaluated with 10 repetitions of 10-fold cross-validation with the percent metric correct. Statistical tests have been performed to find statistically significant differences between the methods, as well as ranking wins-losses of the methods. The results of these tests place \textit{Pairwise Correlation} and \textit{Pairwise Consistency} in the first two positions in the ranking, outperforming univariate and multivariate feature ranking methods and attribute subset evaluation feature selection methods.

The \textit{Pairwise Correlation} and \textit{Pairwise Consistency} methods will be officially published shortly on the \textit{Weka} platform. We are currently analysing the best genes selected with the \textit{Pairwise Correlation} and \textit{Pairwise Consistency} methods for biological interpretation.

\section*{Acknowledgements}
This work was partially funded by the SITSUS project (Ref: RTI2018-094832-B-I00), given by the Spanish Ministry of Science, Innovation and Universities (MCIU), the Spanish Agency for Research (AEI) and by the European Fund for Regional Development (FEDER).
This work was supported by the Science and Technology Agency, S\'eneca Foundation, Comunidad Aut\'onoma Regi\'on de Murcia, Spain through the research projects 00004/COVI/20 and 00007/COVI/20.





\bibliographystyle{elsarticle-num}
\bibliography{references,FSmicroarray}


\appendix
\label{Appendix}
\section{Percent correct evaluation}
\begin{table}[!htpb]
	\resizebox{\textwidth}{!}{\begin{tabular}{ccccccccc}\hline		
			\textbf{\textit{Method}} & 
			\textbf{\textit{NaiveBayes}} &
			\textbf{\textit{MLP}}  &
			\textbf{\textit{SVM}} & \textbf{\textit{kNN}}  &
			\textbf{\textit{RIPPER}}  &
			\textbf{\textit{C4.5}}  &
			\textbf{\textit{Random Forest}}  &
			\textbf{\textit{ZeroR}}\\\hline
			\textit{Clustering Variation}  & 22.27 &   37.15 &   36.84 &   36.63 &   37.10 &   37.45 &   36.97 &   37.45  \\
			\textit{Chi Squared} &  91.16 &   86.50 &   88.53 &   91.90  &   90.53 &    90.97  &   \textbf{93.40} &   37.45  \\
			\textit{Correlation}  & 87.34 &   75.57 &   88.99 &    87.08  &   89.23 &    88.91  &   88.99 &    37.45  \\
			\textit{Information Gain}  & 91.16 &   86.43 &   88.53 &   91.90  &   90.82   &  90.94  &   93.28 &   37.45  \\
			\textit{ReliefF} &  91.30 &   81.38 &   87.93 &   89.44  &   90.19   &  90.29   &  91.88  &   37.45  \\
			\textit{Significance} &  91.16 &   86.54 &   88.50 &   91.90  &   90.52  &   90.82  &   93.31 &   37.45  \\
			\textit{Pairwise Correlation} &  92.57 &   89.64 &   91.24  &   92.64  &   91.11 &    91.97   &  93.36  &   37.45  \\
			\textit{Pairwise Consistency} & 92.66 &   88.99 &   90.85  &   89.93 &   89.28 &   91.85  &   92.47  &   37.45  \\
			\hline			
		\end{tabular}}
		\caption{Average performance evaluation with the gene expression cancer RNA-Seq dataset for $q=3$, 10-fold cross-validation, 10 repetitions.}
		\label{tab:PECancer-q3}
	\end{table}

	\begin{table}[!htpb]
		\resizebox{\textwidth}{!}{\begin{tabular}{ccccccccc}\hline		
				\textbf{\textit{Method}} & 
				\textbf{\textit{NaiveBayes}} &
				\textbf{\textit{MLP}}  &
				\textbf{\textit{SVM}} & \textbf{\textit{kNN}}  &
				\textbf{\textit{RIPPER}}  &
				\textbf{\textit{C4.5}}  &
				\textbf{\textit{Random Forest}}  &
				\textbf{\textit{ZeroR}}\\\hline
				\textit{Clustering Variation}  &   19.84 &   37.37 &   38.08 &   35.79 &   37.34 &   37.45 &   35.94 &   37.45  \\
				\textit{Chi Squared} &   98.99 &   93.17 &   99.16 &    99.10  &   95.78 &   96.47 &   99.21 &    37.45  \\
				\textit{Correlation}  &    97.07 &   86.82 &   98.14 &    97.08 &     92.61 &   93.56 &   97.37  &   37.45  \\
				\textit{Information Gain}  & 99.10 &   92.83 &   99.48   &  99.25  &   95.76 &   96.12 &   99.01 &    37.45  \\
				\textit{ReliefF} &   96.73 &   89.71 &   97.84 &  97.50 &    94.92 &   95.84  &   97.99 &   37.45  \\
				\textit{Significance} &   98.43 &   92.49 &   99.49 &   99.25 &    96.24 &   96.43 &   99.10  &   37.45  \\
				\textit{Pairwise Correlation} &  99.08 &   93.26 &   99.71   &  \textbf{99.75} &    96.24 &   97.05 &   99.18  &   37.45  \\
				\textit{Pairwise Consistency} &  99.01 &   92.48 &   99.20   &  99.25 &    95.18 &   96.22 &   99.18  &   37.45  \\
				\hline			
			\end{tabular}}
			\caption{Average performance evaluation with the gene expression cancer RNA-Seq dataset for $q=\log_2(n)=14$, 10-fold cross-validation, 10 repetitions.}
			\label{tab:PECancer-q14}
		\end{table}

		\begin{table}[!htpb]
			\resizebox{\textwidth}{!}{\begin{tabular}{ccccccccc}\hline		
					\textbf{\textit{Method}} & 
					\textbf{\textit{NaiveBayes}} &
					\textbf{\textit{MLP}}  &
					\textbf{\textit{SVM}} & \textbf{\textit{kNN}}  &
					\textbf{\textit{RIPPER}}  &
					\textbf{\textit{C4.5}}  &
					\textbf{\textit{Random Forest}}  &
					\textbf{\textit{ZeroR}}\\\hline
					\textit{Clustering Variation}   & 26.90 & 42.27 & 41.66 & 39.15 & 43.43 & 43.13 & 41.65 & 37.45 \\
					\textit{Chi Squared}  & 99.60 & 97.29 & 99.84 & 99.85 & 97.42 & 97.69 & 99.73 & 37.45 \\
					\textit{Correlation}   & 98.84 & 96.41 & 99.63 & 99.64 & 95.96 & 95.77 & 99.25 & 37.45 \\
					\textit{Information Gain}   & 99.74 & 97.21 & \textbf{99.88} & \textbf{99.88} & 97.03 & 97.64 & 99.69 & 37.45 \\
					\textit{ReliefF}  & 98.86 & 98.01 & 99.73 & 99.48 & 97.28 & 97.99 & 99.55 & 37.45 \\
					\textit{Significance}  & 98.56 & 96.90 & 99.61 & 99.64 & 96.77 & 97.72 & 99.50 & 37.45 \\
					\textit{Pairwise Correlation}  & 98.63 & 97.33 & 99.86 & 99.85 & 97.53 & 97.40 & 99.51 & 37.45 \\
					\textit{Pairwise Consistency}  & 99.56 & 97.57 & 99.75 & \textbf{99.88} & 97.12 & 97.38 & 99.64 & 37.45 \\
					\hline			
				\end{tabular}}
				\caption{Average performance evaluation with the gene expression cancer RNA-Seq dataset for $q=50$, 10-fold cross-validation, 10 repetitions.}
				\label{tab:PECancer-q50}
			\end{table}	
			
			\begin{table}[!htpb]
				\resizebox{\textwidth}{!}{\begin{tabular}{ccccccccc}\hline		
						\textbf{\textit{Method}} & 
						\textbf{\textit{NaiveBayes}} &
						\textbf{\textit{MLP}}  &
						\textbf{\textit{SVM}} & \textbf{\textit{kNN}}  &
						\textbf{\textit{RIPPER}}  &
						\textbf{\textit{C4.5}}  &
						\textbf{\textit{Random Forest}}  &
						\textbf{\textit{ZeroR}}\\\hline
						\textit{Clustering Variation}   & 40.82 & 45.40 & 44.81 & 41.50 & 47.07 & 49.25 & 48.70 & 37.45 \\
						\textit{Chi Squared}  & 99.61 & 97.29 & 99.76 & 99.75 & 97.67 & 97.75 & 99.61 & 37.45 \\
						\textit{Correlation}   & 99.50 & 95.38 & 99.88 & 99.85 & 96.25 & 96.37 & 99.54 & 37.45 \\
						\textit{Information Gain}   & 99.79 & 98.06 & \textbf{100.00} & 99.88 & 97.08 & 97.43 & 99.78 & 37.45 \\
						\textit{ReliefF}  & 99.36 & 97.78 & 99.90 & \textbf{100.00} & 97.59 & 97.63 & 99.54 & 37.45 \\
						\textit{Significance}  & 99.08 & 98.10 & \textbf{100.00} & 99.75 & 97.77 & 97.29 & 99.64 & 37.45 \\
						\textit{Pairwise Correlation}  & 99.44 & 98.30 & \textbf{100.00} & \textbf{100.00} & 97.53 & 97.90 & 99.65 & 37.45 \\
						\textit{Pairwise Consistency}  & 99.79 & 97.81 & \textbf{100.00} & 99.78 & 97.09 & 97.59 & 99.76 & 37.45 \\
						\hline			
					\end{tabular}}
					\caption{Average performance evaluation with the gene expression cancer RNA-Seq dataset for $q=100$, 10-fold cross-validation, 10 repetitions.}
					\label{tab:PECancer-q100}
				\end{table}

				\begin{table}[!htpb]
					\resizebox{\textwidth}{!}{\begin{tabular}{ccccccccc}\hline		
							\textbf{\textit{Method}} & 
							\textbf{\textit{NaiveBayes}} &
							\textbf{\textit{MLP}}  &
							\textbf{\textit{SVM}} & \textbf{\textit{kNN}}  &
							\textbf{\textit{RIPPER}}  &
							\textbf{\textit{C4.5}}  &
							\textbf{\textit{Random Forest}}  &
							\textbf{\textit{ZeroR}}\\\hline
							\textit{Clustering Variation}   & 15.30 & 11.62 & 11.31 & 17.91 & 14.01 & 17.52 & 19.54 & 11.31 \\
							\textit{Chi Squared}  & 39.77 & 13.02 & 11.31 & 36.42 & 27.20 & 35.50 & 39.21 & 11.31 \\
							\textit{Correlation}   & 18.00 & 11.87 & 10.71 & 15.57 & 13.56 & 16.08 & 17.14 & 11.31 \\
							\textit{Information Gain}   & \textbf{45.55} & 13.09 & 11.37 & 38.13 & 34.17 & 41.27 & 44.64 & 11.31 \\
							\textit{ReliefF}  & 15.38 & 11.70 & 10.73 & 15.93 & 11.52 & 16.40 & 18.93 & 11.31 \\
							\textit{Significance}  & 28.67 & 12.59 & 11.31 & 24.34 & 18.04 & 25.30 & 27.67 & 11.31 \\
							\textit{Pairwise Correlation}  & \textbf{45.55} & 13.09 & 11.37 & 38.13 & 34.17 & 41.27 & 44.64 & 11.31 \\
							\textit{Pairwise Consistency}  & 39.77 & 13.02 & 11.31 & 36.42 & 27.20 & 35.50 & 39.21 & 11.31 \\
							
							\hline			
						\end{tabular}}
						\caption{Average performance evaluation with the brain tissue GTEx RNA expression classification problem for $q=3$, 10-fold cross-validation, 10 repetitions.}
						\label{tab:PEtissue-q3}
					\end{table}	
					
					\begin{table}[!htpb]
						\resizebox{\textwidth}{!}{\begin{tabular}{ccccccccc}\hline		
								\textbf{\textit{Method}} & 
								\textbf{\textit{NaiveBayes}} &
								\textbf{\textit{MLP}}  &
								\textbf{\textit{SVM}} & \textbf{\textit{kNN}}  &
								\textbf{\textit{RIPPER}}  &
								\textbf{\textit{C4.5}}  &
								\textbf{\textit{Random Forest}}  &
								\textbf{\textit{ZeroR}}\\\hline
								\textit{Clustering Variation}   & 39.76 & 12.59 & 10.81 & 24.62 & 23.26 & 28.16 & 43.00 & 11.31 \\
								\textit{Chi Squared}  & 58.14 & 17.15 & 11.31 & 36.92 & 36.21 & 44.40 & 56.50 & 11.31 \\
								\textit{Correlation}   & 30.95 & 14.44 & 8.94 & 21.52 & 14.49 & 23.61 & 30.33 & 11.31 \\
								\textit{Information Gain}   & 59.68 & 17.30 & 11.31 & 40.63 & 39.57 & 45.70 & 59.66 & 11.31 \\
								\textit{ReliefF}  & 28.33 & 14.09 & 9.80 & 28.92 & 20.33 & 24.17 & 36.15 & 11.31 \\
								\textit{Significance}  & 56.76 & 16.88 & 10.18 & 41.13 & 37.14 & 48.63 & 55.95 & 11.31 \\
								\textit{Pairwise Correlation}  & \textbf{66.78} & 17.35 & 10.67 & 41.97 & 48.78 & 54.91 & 65.44 & 11.31 \\
								\textit{Pairwise Consistency}  & 57.20 & 17.45 & 10.92 & 37.53 & 36.71 & 43.42 & 56.57 & 11.31 \\			
								\hline			
							\end{tabular}}
							\caption{Average performance evaluation with the brain tissue GTEx RNA expression classification problem for $q=\log_2(n)=14$, 10-fold cross-validation, 10 repetitions.}
							\label{tab:PEtissue-q14}
						\end{table}	
						
						\begin{table}[!htpb]
							\resizebox{\textwidth}{!}{\begin{tabular}{ccccccccc}\hline		
									\textbf{\textit{Method}} & 
									\textbf{\textit{NaiveBayes}} &
									\textbf{\textit{MLP}}  &
									\textbf{\textit{SVM}} & \textbf{\textit{kNN}}  &
									\textbf{\textit{RIPPER}}  &
									\textbf{\textit{C4.5}}  &
									\textbf{\textit{Random Forest}}  &
									\textbf{\textit{ZeroR}}\\\hline
									\textit{Clustering Variation}   & 55.42 & 12.69 & 9.21 & 25.56 & 32.64 & 38.63 & 58.54 & 11.31 \\
									\textit{Chi Squared}  & 69.21 & 16.13 & 9.07 & 41.73 & 48.55 & 57.40 & 69.20 & 11.31 \\
									\textit{Correlation}   & 43.86 & 13.41 & 8.12 & 27.33 & 18.21 & 26.62 & 41.66 & 11.31 \\
									\textit{Information Gain}   & 68.82 & 16.17 & 9.14 & 42.28 & 48.67 & 56.50 & 68.48 & 11.31 \\
									\textit{ReliefF}  & 35.57 & 14.36 & 7.13 & 33.65 & 22.30 & 26.69 & 43.20 & 11.31 \\
									\textit{Significance}  & 61.37 & 15.07 & 9.12 & 37.86 & 38.07 & 50.45 & 59.66 & 11.31 \\
									\textit{Pairwise Correlation}  & 68.65 & 15.98 & 8.36 & 39.93 & 46.80 & 55.76 & 68.10 & 11.31 \\
									\textit{Pairwise Consistency}  & \textbf{69.35} & 16.66 & 9.57 & 43.52 & 48.22 & 56.17 & 68.63 & 11.31 \\	
									\hline			
								\end{tabular}}
								\caption{Average performance evaluation with the brain tissue GTEx RNA expression classification problem for $q=50$, 10-fold cross-validation, 10 repetitions.}
								\label{tab:PEtissue-q50}
							\end{table}	
							
							\begin{table}[!htpb]
								\resizebox{\textwidth}{!}{\begin{tabular}{ccccccccc}\hline		
										\textbf{\textit{Method}} & 
										\textbf{\textit{NaiveBayes}} &
										\textbf{\textit{MLP}}  &
										\textbf{\textit{SVM}} & \textbf{\textit{kNN}}  &
										\textbf{\textit{RIPPER}}  &
										\textbf{\textit{C4.5}}  &
										\textbf{\textit{Random Forest}}  &
										\textbf{\textit{ZeroR}}\\\hline
										\textit{Clustering Variation}   & 62.55 & 11.73 & 8.71 & 27.74 & 39.90 & 48.90 & 65.44 & 11.31 \\
										\textit{Chi Squared}  & 69.26 & 15.45 & 7.85 & 42.73 & 48.81 & 55.22 & 68.86 & 11.31 \\
										\textit{Correlation}   & 50.72 & 12.52 & 6.70 & 29.28 & 23.09 & 30.94 & 46.05 & 11.31 \\
										\textit{Information Gain}   & 69.52 & 14.88 & 7.14 & 40.88 & 46.53 & 55.83 & 68.24 & 11.31 \\
										\textit{ReliefF}  & 39.75 & 14.34 & 5.68 & 34.06 & 24.56 & 31.99 & 46.41 & 11.31 \\
										\textit{Significance}  & 63.89 & 14.34 & 6.67 & 37.65 & 45.41 & 52.88 & 61.11 & 11.31 \\
										\textit{Pairwise Correlation}  & 68.88 & 14.54 & 6.92 & 40.48 & 45.68 & 54.95 & 67.34 & 11.31 \\
										\textit{Pairwise Consistency}  & \textbf{69.60} & 16.01 & 7.78 & 45.32 & 47.55 & 57.18 & 68.60 & 11.31 \\			
										\hline			
									\end{tabular}}
									\caption{Average performance evaluation with the brain tissue GTEx RNA expression classification problem for $q=100$, 10-fold cross-validation, 10 repetitions.}
									\label{tab:PEtissue-q100}
								\end{table}		
								
								\begin{table}[!htpb]
									\resizebox{\textwidth}{!}{\begin{tabular}{ccccccccc}\hline		
											\textbf{\textit{Method}} & 
											\textbf{\textit{NaiveBayes}} &
											\textbf{\textit{MLP}}  &
											\textbf{\textit{SVM}} & \textbf{\textit{kNN}}  &
											\textbf{\textit{RIPPER}}  &
											\textbf{\textit{C4.5}}  &
											\textbf{\textit{Random Forest}}  &
											\textbf{\textit{ZeroR}}\\\hline
											\textit{Clustering Variation}   & 43.81 & 46.10 & 46.11 & 36.51 & 45.72 & 44.64 & 41.56 & 46.11 \\
											\textit{Chi Squared}  & 45.07 & 46.01 & 46.11 & 40.65 & 46.25 & 42.83 & 43.43 & 46.11 \\
											\textit{Correlation}   & 44.34 & 46.11 & 46.11 & 35.90 & 45.83 & 44.22 & 40.54 & 46.11 \\
											\textit{Information Gain}   & 45.07 & 46.07 & 46.11 & 40.65 & 46.43 & 42.83 & 43.58 & 46.11 \\
											\textit{ReliefF}  & 46.00 & 46.11 & 46.11 & 38.44 & 45.63 & 44.22 & 42.22 & 46.11 \\
											\textit{Significance}  & 47.07 & 46.16 & 46.11 & 39.99 & 47.72 & 45.21 & 44.46 & 46.11 \\
											\textit{Pairwise Correlation}  & 45.07 & 45.88 & 46.11 & 40.65 & 46.04 & 42.88 & 43.67 & 46.11 \\
											\textit{Pairwise Consistency}  & \textbf{48.25} & 46.17 & 46.11 & 41.84 & 47.50 & 45.93 & 46.51 & 46.11 \\			
											\hline			
										\end{tabular}}
										\caption{Average performance evaluation with the brain age GTEx RNA expression classification problem for $q=3$, 10-fold cross-validation, 10 repetitions.}
										\label{tab:PEage-q3}
									\end{table}	
									
									\begin{table}[!htpb]
										\resizebox{\textwidth}{!}{\begin{tabular}{ccccccccc}\hline		
												\textbf{\textit{Method}} & 
												\textbf{\textit{NaiveBayes}} &
												\textbf{\textit{MLP}}  &
												\textbf{\textit{SVM}} & \textbf{\textit{kNN}}  &
												\textbf{\textit{RIPPER}}  &
												\textbf{\textit{C4.5}}  &
												\textbf{\textit{Random Forest}}  &
												\textbf{\textit{ZeroR}}\\\hline
												\textit{Clustering Variation}   & 25.43 & 44.71 & 46.11 & 36.94 & 46.38 & 38.89 & 45.15 & 46.11 \\
												\textit{Chi Squared}  & 35.60 & 44.74 & 46.11 & 44.91 & 47.19 & 44.25 & 49.71 & 46.11 \\
												\textit{Correlation}   & 23.96 & 44.51 & 46.11 & 36.47 & 45.66 & 39.29 & 43.92 & 46.11 \\
												\textit{Information Gain}   & 35.60 & 44.77 & 46.11 & 44.91 & 47.38 & 44.17 & 49.62 & 46.11 \\
												\textit{ReliefF}  & 23.73 & 44.50 & 46.11 & 38.15 & 45.70 & 38.08 & 45.14 & 46.11 \\
												\textit{Significance}  & 35.60 & 45.07 & 46.11 & 44.91 & 47.15 & 44.12 & \textbf{49.73} & 46.11 \\
												\textit{Pairwise Correlation}  & 35.60 & 44.68 & 46.11 & 44.91 & 47.19 & 44.19 & 49.56 & 46.11 \\
												\textit{Pairwise Consistency}  & 35.73 & 44.79 & 46.11 & 43.30 & 47.12 & 43.85 & 49.69 & 46.11 \\
												\hline			
											\end{tabular}}
											\caption{Average performance evaluation with the brain age GTEx RNA expression classification problem for $q=\log_2(n)=14$, 10-fold cross-validation, 10 repetitions.}
											\label{tab:PEage-q14}
										\end{table}	
										
										\begin{table}[!htpb]
											\resizebox{\textwidth}{!}{\begin{tabular}{ccccccccc}\hline		
													\textbf{\textit{Method}} & 
													\textbf{\textit{NaiveBayes}} &
													\textbf{\textit{MLP}}  &
													\textbf{\textit{SVM}} & \textbf{\textit{kNN}}  &
													\textbf{\textit{RIPPER}}  &
													\textbf{\textit{C4.5}}  &
													\textbf{\textit{Random Forest}}  &
													\textbf{\textit{ZeroR}}\\\hline
													\textit{Clustering Variation}   & 21.38 & 42.25 & 46.11 & 36.81 & 46.23 & 38.67 & 44.93 & 46.11 \\
													\textit{Chi Squared}  & 30.68 & 42.93 & 46.11 & 39.72 & 46.39 & 43.06 & 47.04 & 46.11 \\
													\textit{Correlation}   & 18.42 & 42.39 & 46.11 & 36.78 & 45.45 & 37.17 & 44.25 & 46.11 \\
													\textit{Information Gain}   & 30.68 & 42.66 & 46.11 & 39.72 & 46.60 & 43.00 & 47.20 & 46.11 \\
													\textit{ReliefF}  & 21.05 & 42.99 & 46.11 & 39.01 & 45.32 & 37.22 & 44.68 & 46.11 \\
													\textit{Significance}  & 30.68 & 42.53 & 46.11 & 39.72 & 46.66 & 43.06 & 46.86 & 46.11 \\
													\textit{Pairwise Correlation}  & 30.68 & 42.75 & 46.11 & 39.72 & 46.48 & 43.02 & \textbf{47.50} & 46.11 \\
													\textit{Pairwise Consistency}  & 29.64 & 42.97 & 46.11 & 38.98 & 46.59 & 42.07 & 46.62 & 46.11 \\
													\hline			
												\end{tabular}}
												\caption{Average performance evaluation with the brain age GTEx RNA expression classification problem for $q=50$, 10-fold cross-validation, 10 repetitions.}
												\label{tab:PEage-q50}
											\end{table}	
											
								\begin{table}[!htpb]
												\resizebox{\textwidth}{!}{\begin{tabular}{ccccccccc}\hline		
														\textbf{\textit{Method}} & 
														\textbf{\textit{NaiveBayes}} &
														\textbf{\textit{MLP}}  &
														\textbf{\textit{SVM}} & \textbf{\textit{kNN}}  &
														\textbf{\textit{RIPPER}}  &
														\textbf{\textit{C4.5}}  &
														\textbf{\textit{Random Forest}}  &
														\textbf{\textit{ZeroR}}\\\hline
														\textit{Clustering Variation}   & 20.03 & 40.32 & 46.11 & 38.49 & 45.66 & 38.42 & 44.93 & 46.11 \\
														\textit{Chi Squared}  & 28.11 & 41.75 & 46.11 & 37.65 & 46.44 & 41.79 & 46.04 & 46.11 \\
														\textit{Correlation}   & 19.02 & 41.05 & 46.11 & 37.84 & 45.18 & 37.03 & 44.24 & 46.11 \\
														\textit{Information Gain}   & 28.11 & 41.20 & 46.11 & 37.65 & 46.40 & 41.80 & 46.07 & 46.11 \\
														\textit{ReliefF}  & 20.85 & 42.71 & 46.11 & 38.52 & 45.22 & 37.39 & 44.45 & 46.11 \\
														\textit{Significance}  & 28.11 & 41.26 & 46.11 & 37.65 & \textbf{46.60} & 41.80 & 46.25 & 46.11 \\
														\textit{Pairwise Correlation}  & 28.11 & 41.07 & 46.11 & 37.65 & 46.42 & 41.78 & 46.02 & 46.11 \\
														\textit{Pairwise Consistency}  & 27.36 & 41.62 & 46.11 & 37.76 & 46.48 & 41.51 & 46.00 & 46.11 \\			
														\hline			
													\end{tabular}}
													\caption{Average performance evaluation with the brain age GTEx RNA expression classification problem for $q=100$, 10-fold cross-validation, 10 repetitions.}
													\label{tab:PEAGE-q100}
												\end{table}

\begin{table}[!htpb]
	\resizebox{\textwidth}{!}{\begin{tabular}{cccccccccc}\hline		
			\textbf{\textit{Method}} & \textbf{\textit{Attributes}} &
			\textbf{\textit{NaiveBayes}} &
			\textbf{\textit{MLP}}  &
			\textbf{\textit{SVM}} & \textbf{\textit{kNN}}  &
			\textbf{\textit{RIPPER}}  &
			\textbf{\textit{C4.5}}  &
			\textbf{\textit{Random Forest}}  &
			\textbf{\textit{ZeroR}}\\\hline
\textit{MOEA-CFS} & 141 & 99.65 & 96.17 & 99.99 & 99.88 & 98.00 & 97.62 & 99.65 & 37.45 \\
\textit{MOEA-Consistency} & 25 & 84.23 & 72.98 & 88.56 & 85.27 & 77.25 & 73.98 & 87.60 & 37.45 \\
\textit{Pairwise Correlation ($ q=100 $)} & 100 & 99.44 & 98.30 & \textbf{100.00} & \textbf{100.00} & 97.53 & 97.90 & 99.65 & 37.45 \\
\textit{Pairwise Consistency ($ q=100 $)} & 100 & 99.79 & 97.81 & \textbf{100.00} & 99.78 & 97.09 & 97.59 & 99.76 & 37.45 \\
\textit{Pairwise Correlation ($ q=200 $)} & 200 & 99.99 & 98.01 & 99.96 & 99.78 & 97.94 & 98.12 & 99.65 & 37.45 \\
\textit{Pairwise Consistency ($ q=200 $)} & 200 & 99.55 & 97.58 & \textbf{100.00} & 99.75 & 97.23 & 97.23 & 99.76 & 37.45 \\
			\hline			
		\end{tabular}}
		\caption{Average performance evaluation with gene expression cancer RNA-Seq dataset, 10-fold cross-validation, 10 repetitions.}
		\label{tab:PECancer-SE}
\end{table}	

\begin{table}[!htpb]
	\resizebox{\textwidth}{!}{\begin{tabular}{cccccccccc}\hline		
			\textbf{\textit{Method}} & \textbf{\textit{Attributes}} & 
			\textbf{\textit{NaiveBayes}} &
			\textbf{\textit{MLP}}  &
			\textbf{\textit{SVM}} & \textbf{\textit{kNN}}  &
			\textbf{\textit{RIPPER}}  &
			\textbf{\textit{C4.5}}  &
			\textbf{\textit{Random Forest}}  &
			\textbf{\textit{ZeroR}}\\\hline
\textit{MOEA-CFS} & 123 & \textbf{70.37} & 13.90 & 6.66 & 38.83 & 47.62 & 56.64 & 68.52 & 11.31 \\
\textit{MOEA-Consistency} & 89 & 52.75 & 12.34 & 7.57 & 27.38 & 27.70 & 36.25 & 50.09 & 11.31 \\
\textit{Pairwise Correlation ($ q=100 $)} & 100 & 68.88 & 14.54 & 6.92 & 40.48 & 45.68 & 54.95 & 67.34 & 11.31 \\
\textit{Pairwise Consistency ($ q=100 $)} & 100 & 69.60 & 16.01 & 7.78 & 45.32 & 47.55 & 57.18 & 68.60 & 11.31 \\
\textit{Pairwise Correlation ($ q=200 $)} & 200 & 69.54 & 14.39 & 3.99 & 39.93 & 45.79 & 54.03 & 66.54 & 11.31 \\
\textit{Pairwise Consistency ($ q=200 $)} & 200 & 70.15 & 15.08 & 5.36 & 45.95 & 47.85 & 56.48 & 67.84 & 11.31 \\			
			\hline			
		\end{tabular}}
		\caption{Average performance evaluation with brain tissue GTEx RNA expression classification problem, 10-fold cross-validation, 10 repetitions.}
		\label{tab:PEtissue-SE}
	\end{table}	

\begin{table}[!htpb]
	\resizebox{\textwidth}{!}{\begin{tabular}{cccccccccc}\hline		
			\textbf{\textit{Method}} & \textbf{\textit{Attributes}} &
			\textbf{\textit{NaiveBayes}} &
			\textbf{\textit{MLP}}  &
			\textbf{\textit{SVM}} & \textbf{\textit{kNN}}  &
			\textbf{\textit{RIPPER}}  &
			\textbf{\textit{C4.5}}  &
			\textbf{\textit{Random Forest}}  &
			\textbf{\textit{ZeroR}}\\\hline
\textit{MOEA-CFS} & 135 & 23.32 & 39.96 & 46.11 & 39.17 & 45.65 & 38.95 & 44.76 & 46.11 \\
\textit{MOEA-Consistency} & 117 & 26.39 & 40.41 & 46.11 & 38.80 & 46.33 & 40.61 & 45.98 & 46.11 \\
\textit{Pairwise Correlation ($ q=100 $)} & 100 & 28.11 & 41.07 & 46.11 & 37.65 & 46.42 & 41.78 & 46.02 & 46.11 \\
\textit{Pairwise Consistency ($ q=100 $)} & 100 & 27.36 & 41.62 & 46.11 & 37.76 & \textbf{46.48} & 41.51 & 46.00 & 46.11 \\
\textit{Pairwise Correlation ($ q=200 $)} & 200 & 25.46 & 39.39 & 46.11 & 37.76 & 45.89 & 40.96 & 45.12 & 46.11 \\
\textit{Pairwise Consistency ($ q=200 $)} & 200 & 24.81 & 39.67 & 46.11 & 37.87 & 46.08 & 40.97 & 45.20 & 46.11 \\			
			\hline			
		\end{tabular}}
		\caption{Average performance evaluation  with brain age GTEx RNA expression classification problem, 10-fold cross-validation, 10 repetitions.}
		\label{tab:PEage-SE}
	\end{table}	

\clearpage
\section{Statistical tests: wins-losses ranking}

\begin{table}[!htpb]
	\resizebox{\textwidth}{!}{\begin{tabular}{c|ccc|ccc|ccc|ccc|}\cline{2-13}

			& \multicolumn{3}{c|}{$q=3$}
			& \multicolumn{3}{c|}{$q=\log_2(n)=14$}
			& \multicolumn{3}{c|}{$q=50$}
			& \multicolumn{3}{c|}{$q=100$}
			\\ \hline
			
			\multicolumn{1}{|c|}{\textbf{\textit{Method}}} & \textbf{\textit{wins}} & \textbf{\textit{losses}} & \textbf{\textit{dif.}} &  \textbf{\textit{wins}} & \textbf{\textit{losses}} & \textbf{\textit{dif.}} & \textbf{\textit{wins}} & \textbf{\textit{losses}} & \textbf{\textit{dif.}}  & \textbf{\textit{wins}} & \textbf{\textit{losses}} & \textbf{\textit{dif.}}  \\\hline
			\multicolumn{1}{|c|}{{\textit{Clustering Variation}}}   & 0 & 49 & -49 & 0 & 49 & -49 & 0 &  49 & -49 & 0 & 49 & -49  \\
			\multicolumn{1}{|c|}{{\textit{Chi Squared}}}   & 12 & 4 & 8 & 17 & 1 & 16 & 10 &  0 & 10 & 7 & 0 & 7  \\
			\multicolumn{1}{|c|}{{\textit{Correlation}}}   & 7 & 24 & -17 & 7 & 33 & -26 & 7 & 7 & 0 & 7 & 1 & 6  \\
			\multicolumn{1}{|c|}{{\textit{Information Gain}}}   & 12 & 4 & 8 & 17 & 0 & 17 & 12 & 0 & \textbf{12} & 8 & 0 & \textbf{8} \\
			\multicolumn{1}{|c|}{{\textit{ReliefF}}}   & 10 & 8 & 2 & 8 & 18 & -10 & 8 & 1 & 7 & 7 & 0 & 7 \\
			\multicolumn{1}{|c|}{{\textit{Significance}}}   & 12 & 4 & 8 & 17 & 0 & 17 & 8 & 3 & 5 & 7 & 2 & 5 \\
			\multicolumn{1}{|c|}{{\textit{Pairwise Correlation}}}   & 22 & 0 & \textbf{22} & 19 & 0 & \textbf{19} & 8 &  3& 5 & 8 & 0 & \textbf{8} \\
			\multicolumn{1}{|c|}{{\textit{Pairwise Consistency}}}   & 19 & 1 & 18 & 17 & 1 & 16 & 10 & 0 & 10 & 8 & 0 & \textbf{8} \\

			\hline			
		\end{tabular}}
		\caption{Wins $-$ losses ranking tests for gene expression cancer RNA-Seq dataset, 10-fold cross-validation, 10 repetitions.}
		\label{tab:cancer}
	\end{table}

	\begin{table}[!htpb]
		\resizebox{\textwidth}{!}{\begin{tabular}{c|ccc|ccc|ccc|ccc|}\cline{2-13}

				& \multicolumn{3}{c|}{$q=3$}
				& \multicolumn{3}{c|}{$q=\log_2(n)=14$}
				& \multicolumn{3}{c|}{$q=50$}
				& \multicolumn{3}{c|}{$q=100$}
				\\ \hline
				\multicolumn{1}{|c|}{\textbf{\textit{Method}}} & \textbf{\textit{wins}} & \textbf{\textit{losses}} & \textbf{\textit{dif.}} &  \textbf{\textit{wins}} & \textbf{\textit{losses}} & \textbf{\textit{dif.}} & \textbf{\textit{wins}} & \textbf{\textit{losses}} & \textbf{\textit{dif.}}  & \textbf{\textit{wins}} & \textbf{\textit{losses}} & \textbf{\textit{dif.}}  \\\hline
				\multicolumn{1}{|c|}{{\textit{Clustering Variation}}}  & 2 & 26 & -24 & 10 & 31 & -21 & 9 & 30 & -21 & 10 & 30 & -20  \\
				\multicolumn{1}{|c|}{{\textit{Chi Squared}}}   & 20 & 8 & 12 & 19 & 11 & 8 & 22 &  0 & 22 & 22 & 1 & 21  \\
				\multicolumn{1}{|c|}{{\textit{Correlation}}}   & 3 & 26 & -23 & 0 & 41 & -41 & 1 & 35 & -34 & 1 & 35 & -34 \\
				\multicolumn{1}{|c|}{{\textit{Information Gain}}}   & 28 & 0 & \textbf{28} & 29 & 4 & 25 & 23 & 0 & 23 & 21& 1 & 20 \\
				\multicolumn{1}{|c|}{{\textit{ReliefF}}}   & 0 & 28 & -28 & 4 & 34 & -30 & 3 & 33 & -30 & 4 & 32 & -28  \\
				\multicolumn{1}{|c|}{{\textit{Significance}}}   & 15 & 20 & -5 & 21 & 6 & 15 & 15 &  19 & -4 & 14 & 16 & 2 \\
				\multicolumn{1}{|c|}{{\textit{Pairwise Correlation}}}   & 28 & 0 & \textbf{28} & 36 & 0 & \textbf{36} & 21 &  2 & 19 & 20 & 4 & 16 \\
				\multicolumn{1}{|c|}{{\textit{Pairwise Consistency}}}   & 20 & 8 & 12 & 19 & 11 & 8 & 25 &  0 & \textbf{25} & 27 & 0 & \textbf{27} \\
				
				\hline			
			\end{tabular}}
			\caption{Wins $-$ losses ranking tests for brain tissue GTEx RNA expression classification problem, 10-fold cross-validation, 10 repetitions.}
			\label{tab:braintissue}
		\end{table}

		\begin{table}[!htpb]
			\resizebox{\textwidth}{!}{\begin{tabular}{c|ccc|ccc|ccc|ccc|}\cline{2-13}

					& \multicolumn{3}{c|}{$q=3$}
					& \multicolumn{3}{c|}{$q=\log_2(n)=14$}
					& \multicolumn{3}{c|}{$q=50$}
					& \multicolumn{3}{c|}{$q=100$}
					\\ \hline
					\multicolumn{1}{|c|}{\textbf{\textit{Method}}} & \textbf{\textit{wins}} & \textbf{\textit{losses}} & \textbf{\textit{dif.}} &  \textbf{\textit{wins}} & \textbf{\textit{losses}} & \textbf{\textit{dif.}} & \textbf{\textit{wins}} & \textbf{\textit{losses}} & \textbf{\textit{dif.}}  & \textbf{\textit{wins}} & \textbf{\textit{losses}} & \textbf{\textit{dif.}}  \\\hline
					\multicolumn{1}{|c|}{{\textit{Clustering Variation}}}  & 0 & 10 & -10 & 0 & 20 & -20 & 1 & 19 & -18 & 0 & 9 & -9 \\
					\multicolumn{1}{|c|}{{\textit{Chi Squared}}}   & 3 & 3 & 0 & 12 & 0 & \textbf{12} & 11 & 0 & \textbf{11} & 7 & 0 & 7 \\
					\multicolumn{1}{|c|}{{\textit{Correlation}}}  & 0 & 15 & -15 & 0 & 20 & -20 & 0 & 21 & -21 & 0 & 13 & -13 \\
					\multicolumn{1}{|c|}{{\textit{Information Gain}}} & 3 & 3 & 0 & 12 & 0 & \textbf{12} & 11 & 0 & \textbf{11} & 6 & 0 & 6 \\
					\multicolumn{1}{|c|}{{\textit{ReliefF}}}   & 1 & 5 & -4 & 0 & 20 & -20 & 0 & 15 & -14 & 0 & 11 & -11 \\
					\multicolumn{1}{|c|}{{\textit{Significance}}}   & 9 & 0 & 9 & 12 & 0 & \textbf{12} & 11 & 0 & \textbf{11} & 8 & 0 & \textbf{8} \\
					\multicolumn{1}{|c|}{{\textit{Pairwise Correlation}}}   & 3 & 4 & -1 & 12 & 0 & \textbf{12} & 11 & 0 & \textbf{11} & 7 & 0 & 7 \\
					\multicolumn{1}{|c|}{{\textit{Pairwise Consistency}}}   & 21 & 0 & \textbf{21} & 12 & 0 & \textbf{12} & 9 & 0 & 9 & 5 & 0 & 5 \\
					
					\hline			
				\end{tabular}}
				\caption{Wins $-$ losses ranking tests for brain age GTEx RNA expression classification problem, 10-fold cross-validation, 10 repetitions.}
				\label{tab:brainage}
			\end{table}	

		\begin{table}[!htpb]
			\resizebox{\textwidth}{!}{\begin{tabular}{c|ccc|ccc|ccc|}\cline{2-10}

					& \multicolumn{3}{c|}{\textbf{\textit{Cancer RNA-Seq}}}
					& \multicolumn{3}{c|}{\textbf{\textit{Brain tissue GTEx RNA}}}
					& \multicolumn{3}{c|}{\textbf{\textit{Brain age GTEx RNA}}}
					\\ \hline
					\multicolumn{1}{|c|}{\textbf{\textit{Method}}} & \textbf{\textit{wins}} & \textbf{\textit{losses}} & \textbf{\textit{dif.}} &  \textbf{\textit{wins}} & \textbf{\textit{losses}} & \textbf{\textit{dif.}} & \textbf{\textit{wins}} & \textbf{\textit{losses}} & \textbf{\textit{dif.}}    \\\hline
					\multicolumn{1}{|c|}{{\textit{MOEA-CFS}}}  & 7 & 0 & \textbf{7} & 8 & 3 & 5 & 0 & 4 & -4\\
					\multicolumn{1}{|c|}{{\textit{MOEA-Consistency}}} & 0 & 35 & -35 & 2 & 29 & -27 & 1 & 0 & 1\\
					\multicolumn{1}{|c|}{{\textit{Pairwise Correlation ($ q=100 $)}}}& 7 & 0 & \textbf{7} & 7 & 6 & 1 & 3 & 0 & \textbf{3}\\
					\multicolumn{1}{|c|}{{\textit{Pairwise Consistency ($ q=100 $)}}}& 7 & 0 & \textbf{7} & 16 & 0 & \textbf{16} & 3 & 0 & \textbf{3}\\
					\multicolumn{1}{|c|}{{\textit{Pairwise Correlation ($ q=200 $)}}}& 7 & 0 & \textbf{7} & 6 & 9 & -3 & 1 & 2 & -1\\
					\multicolumn{1}{|c|}{{\textit{Pairwise Consistency ($ q=200 $)}}}& 7 & 0 & \textbf{7} & 10 & 2 & 8 & 0 & 2 & -2\\		
					\hline			
				\end{tabular}}
				\caption{Wins $-$ losses ranking tests including attribute subset evaluation feature selection methods, 10-fold cross-validation, 10 repetitions.}
				\label{tab:SEtest}
			\end{table}	

\end{document}